\documentclass[default,iicol,table,xcdraw]{sn-jnl}
\usepackage[T2A]{fontenc}
\usepackage[utf8]{inputenc}
\usepackage{array}
\usepackage[russian,ukrainian,english]{babel}
\usepackage{latexsym}
\usepackage{microtype}
\usepackage{float}
\usepackage{placeins}
\usepackage{graphicx}

\jyear{2021}%

\theoremstyle{thmstyleone}%
\theoremstyle{thmstyletwo}%

\theoremstyle{thmstylethree}%

\begin{document}

\title[Article Title]{Automated multilingual detection of Pro-Kremlin propaganda in newspapers and Telegram posts}


\author*[1]{\fnm{Veronika} \sur{Solopova}}\email{veronika.solopova@fu-berlin.de}

\author[2]{\fnm{Oana-Iuliana} \sur{Popescu}}\email{oana.iuliana.popescu@gmail.com}

\author[1]{\fnm{Christoph} \sur{Benzmüller}}\email{c.benzmueller@fu-berlin.de}
\author[1]{\fnm{Tim} \sur{Landgraf}}\email{tim.landgraf@fu-berlin.de}

\affil*[1]{\orgdiv{Dahlem Center for Machine Learning and Robotics}, \orgname{Freie Universität Berlin}, \country{Germany}}

\affil[2]{\orgname{German Aerospace Center}, \city{Jena}, \country{Germany}}


\abstract{The full-scale conflict between the Russian Federation and Ukraine generated an unprecedented amount of news articles and social media data reflecting opposing ideologies and narratives. These polarized campaigns have led to mutual accusations of misinformation and fake news, shaping an atmosphere of confusion and mistrust for readers worldwide. This study analyses how the media affected and mirrored public opinion during the first month of the war using news articles and Telegram news channels in Ukrainian, Russian, Romanian and English.
We propose and compare two methods of multilingual automated pro-Kremlin propaganda identification, based on Transformers and linguistic features. We analyse the advantages and disadvantages of both methods, their adaptability to new genres and languages, and ethical considerations of their usage for content moderation. With this work, we aim to lay the foundation for further development of moderation tools tailored to the current conflict. }

\maketitle

\section{Introduction}\label{sec1}
Propaganda influences an audience to support a political agenda. \cite{smith2022propaganda,2022propaganda}. Propaganda has been shown to play a vital role in the Russian invasion of Ukraine, shaping the war approval rate \cite{longforwar} by e.g. fabricating explanations for war crimes \cite{war_crimeslies}. As a result, fake news also spreads through Ukrainian, Central European and Western media \cite{propogandawar}, seeding mistrust and confusion \cite{division}.

With every day of the war having a large amount of potentially false information produced, human quality control thereof is limited. Especially during a war, the journalistic virtue of fact-checking may be substantially obstructed. This poses the question of whether statistical analysis can provide us with a reliable prediction of the intent behind a piece of news.
Given a sufficiently high separability, automatic moderation tools could process the news and warn readers about the potential disinformation instead of fully placing the responsibility on human moderators \cite{Steiger2021ThePW}. The objective of this study is to detect war propaganda produced in the context of the 2022 Russian invasion of Ukraine in a transparent, explainable way, as such a tool can be used for content moderation in social media.\\ While Russian propaganda creates a ‘cacophony’ of fabricated opinions and sources in its media ecosystem \cite{nato}, it also has a number of uniform strategies and approaches which are recurrently mentioned in research throughout the whole of Russian-Ukrainian war by international bodies and independent researchers\cite{brown,stef, genocide,gec}. Hence, we hypothesize and aim to prove that propaganda can be successfully detected using certain stylistic and syntactical features behind these strategies, independent of keywords. Naturally, keywords change depending on the course of events, while the tactics of the propaganda stay similar. Traditional algorithms empowered by such features are inherently interpretable and may perform on par with intransparent neural networks. 
Here, we propose a linguistics-based approach for detecting war propaganda in the context of the ongoing war between Ukraine and Russia since the start of the full-scale invasion in 2022, using models trained on news from media outlets from Ukraine, Romania, Great Britain and the USA.
We extracted news from fact-checked outlets identified as credible sources and from outlets recognized as spreading fake news.
We train and evaluate classifiers using a set of linguistic features and keywords, and a multilingual Transformer to classify articles as containing pro-Kremlin and pro-Western narrative indicators.
\\With this work, we provide an open-source tool for the identification of such fake news and propaganda in Russian and Ukrainian, which, to the best of our knowledge, is the first of its kind. We demonstrate that discriminating propaganda from neutral news is not entirely possible in the current situation, as news from both sides may contain war propaganda, and Western media is heavily dependent on Ukrainian official reports. 

In Section \ref{sec:related} we present previous work related to our research. In Section \ref{sec:methods} we first describe the sources of our data and its collection process (3.1-3.2), then, we expand upon the linguistic features (3.3) and the keywords that we extract (3.4). In Section \ref{sec:data} we introduce our training setup for each experiment, describing the data and model configurations, while in Section \ref{sec:results} we present the results for each setting. In Section \ref{sec:analyses} we provide additional analyses, looking into feature importance coefficients of some models (6.1) and distributional exploratory analysis of the classes (6.2), exploring chronological, language and narrative-specific differences. In Section \ref{sec:disussion} we consider the main findings  and limitations of our work. We lay the ground for future work opportunities and delve into ethical dangers in terms of its potential usage for automated content moderation. In Section \ref{sec13} we summarize the main contributions of our study.

\section{Related Work}
\label{sec:related}

Our work is motivated by the fact that despite the assumed involvement of Russia in Brexit \cite{Narayanan2017RussianIA} and the 2016 US presidential elections \cite{Polarize}, there is still only a small number of peer-reviewed research publications investigating Russian state-sponsored fake news \cite{10.1093/joc/jqaa027,ruschin}. Furthermore, existing publications are not always neutral, with some using accusations and emotional lexicon \cite{posttruth}, while others accuse Western media of propaganda used to discredit Russia \cite{ling}.

\citet{aproornot} examines claims of Russian propaganda targeting the US population by analysing weblogs and media sources in terms of their attitude towards Russia and finding a positive correlation between the Russian media outlet Sputnik and several US media sources. 

Timothy Snyder in his books \cite{nla.cat-vn7732877,CRL16744} analyses the Kremlin's informational campaign against the sovereignty of the United States and the integrity of the European Union.

Several studies investigated Russian bots in social media. \citet{troll} used a decision tree classifier on features extracted from the tweets, concluding that bot accounts tend to sound more formal or structured, whereas real user accounts tend to be more informal, containing slang, slurs, and cursing; \citet{ruschin} analyzed the geography and history of the accounts, as well as their market share using Botometer, pointing to a ‘sophisticated and well-resourced campaign by Russia’s Internet Research Agency’. \citet{Narayanan2017RussianIA} performed a basic descriptive analysis to discern how bots were being used to amplify political communication during the Brexit campaign. 
There is a considerable amount of research focusing on fake news detection. \citet{doi:10.1177/2053951719843310} show-cased that significant performance can be achieved even with n-grams; \citet{9089487,Mahyoob2020LinguisticBasedDO,rashkin-etal-2017-truth} implemented different linguistic feature-based solutions, while \citet{https://doi.org/10.48550/arxiv.2101.02359} demonstrated the application of Transformers.
 While fake news makes up a big part of the propaganda toolkit, propaganda also relies on specific wording, appealing to emotions or stereotypes, flag-waving and distraction techniques such as red herrings and whataboutisms \cite{DBLP:journals/corr/abs-2108-12802}.
Research on propaganda detection is less frequent. Although existing works  proposed using both feature-based and contextual embedding approaches \cite{10.1145/3407023.3409211,DBLP:journals/corr/abs-2108-12802,Oliinyk2020PropagandaDI,DBLP:journals/corr/abs-2002-06644}, these studies focused mostly on the English language. To the best of our knowledge, there are no benchmark corpora and no open-source multilingual tools available.

\section{Methods}
\label{sec:methods}
We implement a binary classification using the following models for input vectors consisting of 41 handcrafted linguistic features and 116 keywords (normalized by the length of the text in tokens): decision tree, linear regression, support vector machine (SVM) and neural networks, using stratified 5-fold cross-validation (10\% for test and 90\% for training). For comparison with learned features, we extract embeddings using a multilingual BERT model \cite{DBLP:journals/corr/abs-1810-04805} and train a linear model using them.

\begin{figure}

\includegraphics[width=8cm,height=8cm]{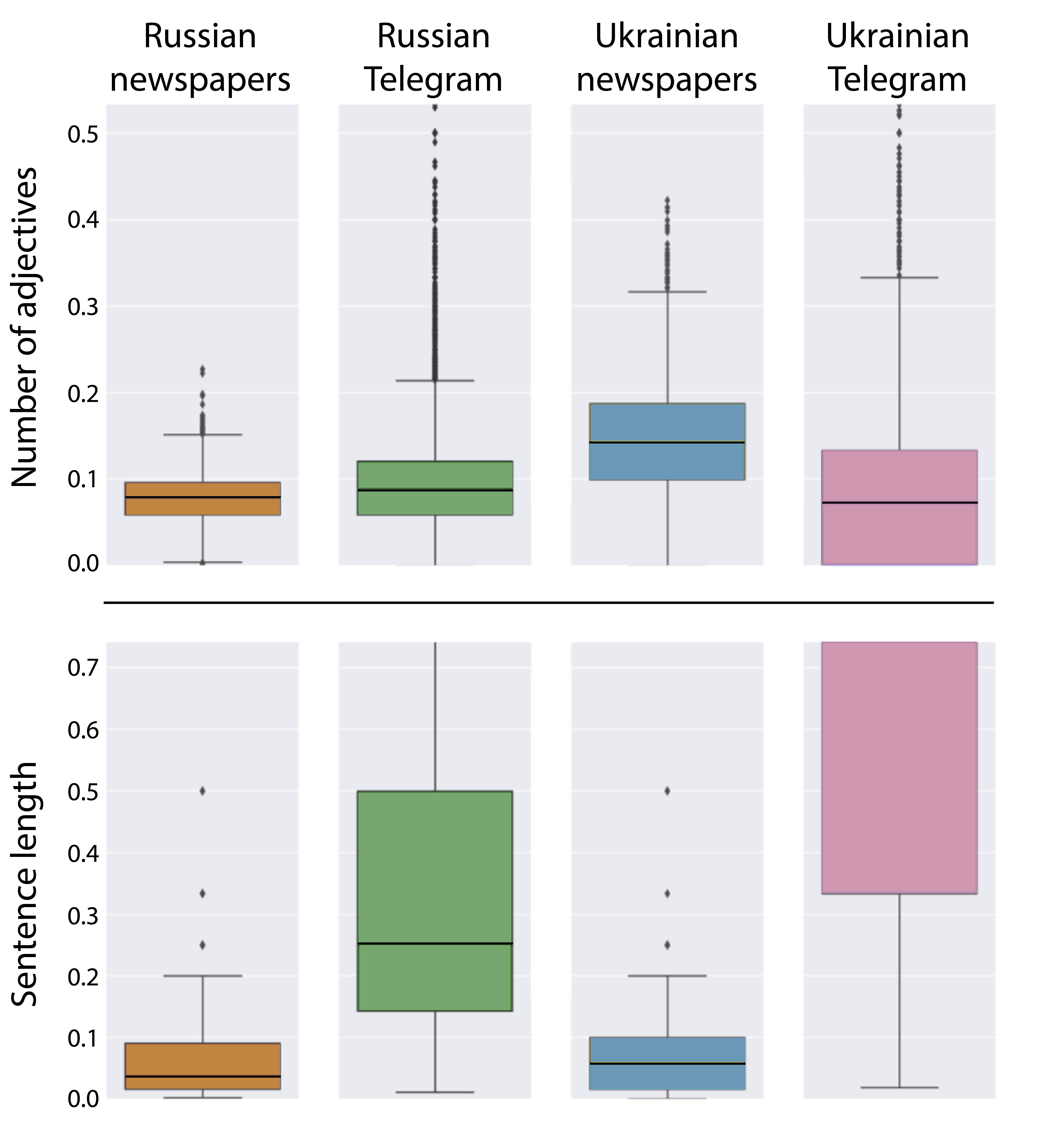}
\caption{Examples of genre-related differences between Newspapers and Telegram subsets. The boxplot represents 25\% around the median, the whiskers show the first and last quartiles. Ukrainian news has much more adjectives than Telegram posts, while it is vice-versa for Russian news. Sentences are much longer in Telegram for both languages. }
\label{diffs}
\end{figure}

We performed 3 sets of experiments contrasting the handcrafted and learned features:\\
\textbf{Experiment 1.} Training models on Russian, Ukrainian, Romanian and English newspaper articles, and evaluating them on the test sets of these languages (1.1) and on French newspaper articles (1.2). We add the French newspapers to benchmark the multilingualism of our models. We choose French because it is in the same language family as Romanian.
\\
\textbf{Experiment 2.} Training models on Russian, Ukrainian, Romanian, English and French newspaper articles, and validating them on the test set (2.1). Additionally, we use this model to test the Russian and Ukrainian Telegram data (2.2.). Here the goal is to investigate whether this model will perform well out-of-the-box for the Telegram articles, which are 10 to 20 times shorter. See an example of the genre-related difference in  distributions in Figure \ref{diffs}.
\\
\textbf{Experiment 3.} Training models on the combined newspaper and Telegram data and applying them to the test set. Here we verify whether adding the Telegram data to the training set can improve generalization power, although data distributions differ.

\begin{table*}[!htbp]

\caption{Corpus statistics, including the sources per language and stance.}
\label{tab:corpus}
\resizebox{\textwidth}{!}{%
\begin{tabular}{|lll|}
\hline
\multicolumn{1}{|c|}{Language} &
  \multicolumn{1}{c|}{Source} &
  Amount of texts \\ \hline
\rowcolor[HTML]{FFFFFF} 
\multicolumn{3}{|c|}{\cellcolor[HTML]{FFFFFF}Pro-Western newspapers} \\ \hline

\multicolumn{1}{|l|}{Ukrainian} &
  \multicolumn{1}{l|}{‘Europeiska Pravda’,‘Ukrainska Pravda’, ‘Espresso’, ‘5.ua’, ‘Hhromadske’, ‘Liga.net’} &
  3298 \\ \hline
\rowcolor[HTML]{FFFFFF} 
\multicolumn{1}{|l|}{\cellcolor[HTML]{FFFFFF}Romanian} &
  \multicolumn{1}{l|}{\cellcolor[HTML]{FFFFFF}‘digi24’, ‘mediafax’, ‘g4media’} &
  \cellcolor[HTML]{FFFFFF}4049 \\ \hline
\multicolumn{1}{|l|}{English} &
  \multicolumn{1}{l|}{‘The Guardian’, ‘BBC’, ‘The New York Times’, ‘Reuters’)} &
  \cellcolor[HTML]{FFFFFF}1060 \\ \hline
\rowcolor[HTML]{FFFFFF} 
\multicolumn{1}{|l|}{\cellcolor[HTML]{FFFFFF}French} &
  \multicolumn{1}{l|}{\cellcolor[HTML]{FFFFFF}‘Tv5monde’, ‘Le Monde’ and ‘Le Figaro’} &
  \cellcolor[HTML]{FFFFFF}458 \\ \hline
\multicolumn{1}{|l|}{Russian} &
  \multicolumn{1}{l|}{‘Raintv’} &
  \cellcolor[HTML]{FFFFFF}7 \\ \hline
\rowcolor[HTML]{FFFFFF} 
\multicolumn{3}{|c|}{\cellcolor[HTML]{FFFFFF}Pro-Kemlin newspapers} \\ \hline
\multicolumn{1}{|l|}{Ukrainian} &
  \multicolumn{1}{l|}{‘Newsua’, ‘Strana.ua’, ‘Vesti.ua’, ‘Ukranews’, ‘Zik’} &
  3579 (474  in Ukrainian  and  3105 in Russian) \\ \hline
\rowcolor[HTML]{FFFFFF} 
\multicolumn{1}{|l|}{\cellcolor[HTML]{FFFFFF}Romanian} &
  \multicolumn{1}{l|}{\cellcolor[HTML]{FFFFFF}‘Antena3’, ‘Stiripesurse’, ‘Romaniatv.net’, ‘Cyd.ro’, ‘Activenews’ and ‘Dcnews’.} &
  3007 \\ \hline
\multicolumn{1}{|l|}{French} &
  \multicolumn{1}{l|}{‘RT’ French edition} &
  123 \\ \hline
\rowcolor[HTML]{FFFFFF} 
\multicolumn{1}{|l|}{\cellcolor[HTML]{FFFFFF}Russian} &
  \multicolumn{1}{l|}{\cellcolor[HTML]{FFFFFF}‘Ria news’, ‘Russia Today’, ‘Interfax’, ‘Lenta.ru’ and ‘Ukraine.ru’.} &
  2648 \\ \hline
  \multicolumn{3}{|c|}{\cellcolor[HTML]{FFFFFF}Telegram posts} \\ \hline
   \rowcolor[HTML]{FFFFFF} 
  \multicolumn{1}{|l|}{Ukrainian} &
  \multicolumn{1}{l|}{‘Goncharenko’, ‘InformNapalm’,‘Brati po zbroi’, ‘Spravdi’,‘Operativni ZSU’} &
 7263 ( 1568  in Ukrainian  and    1568 in Russian) \\ \hline

\multicolumn{1}{|l|}{\cellcolor[HTML]{FFFFFF}Russian} &
  \multicolumn{1}{l|}{\cellcolor[HTML]{FFFFFF}‘Rybar’,‘Siloviki’,‘Vysokigovorit’} &
 61595 \\ \hline
\end{tabular}
}
\end{table*}

\section{Data}
\label{sec:data}

\subsection{Newspapers}

We automatically scraped articles from online news outlets using the  newspaper\footnote{https://newspaper.readthedocs.io/en/latest/} framework. Our data collection spans the period from the 23rd of February, on the eve of the Russian full-scale attack on Ukraine, until the fourth of April, and we sample at eight time points during that period. 

Our choice of media outlets and languages is based on the geopolitical zones which might have been affected by propaganda. We collected news from Ukrainian and Russian media outlets, choosing sources that support pro-Kremlin narratives in Ukraine that have been confirmed by journalistic investigations to directly copy pieces of news from Russian news outlets \cite{prorusmedia}. We included American and British English-speaking outlets as a positive control of widely recognised high-quality news, as well as French news sources. We also added Romanian news as representative of the Central European block, which is one of the secondary targets of propaganda campaigns \cite{posttruth}, and used websites that have been categorized by Rubrika.ro\footnote{https://rubrika.ro/extensie-browser} as containing fake news. Except for English, all languages have two subsets, one supporting the Russian side of the conflict, and one supporting the Ukrainian one. In total, we collected 18,229 texts: 8872 texts featuring pro-Western narratives and 9357 reflecting the official position of the Russian government. The sources are listed in Table \ref{tab:corpus}.

Note that the ground-truth labels were assigned only according to the news source without manual labelling.

\subsection{Telegram posts}

Since the start of the war, many Telegram channels became widely used in Ukraine and Russia for quicker updates on the war and for posting uncensored reports \cite{telegram1,telegram2,telegram3}. However, it is a source without moderation and fact-checking, hence fake news and emotional lexicon, including profanity and propaganda, are not unusual \cite{telegram4}. Therefore, we included both Russian and Ukrainian Telegram news in our data collection.

\subsection{Linguistic Feature Selection}

We start processing the collected texts by extracting per-article linguistic features. The first set of features have been used previously in \cite{Mahyoob2020LinguisticBasedDO} to detect fake news: a number of negations, adverbs, average sentence length, proper nouns, passive voice, quotes, conjunctions (we also count separately the frequency of the conjunction ‘but’ to capture contrasting), comparative and superlative adjectives, state verbs, personal pronouns, modal verbs, interrogatives. Since fake news and propaganda can be associated with ‘overly emotional’ language \cite{doi:10.1177/2053951719843310}, we generate word counts for each basic emotion category: anger, fear, anticipation, trust, surprise, sadness, joy, disgust, and identify two sentiment classes, negative and positive, using the NRC Word-Emotion Association Lexicon \cite{Mohammad13}. Following Rashkin et. al. \cite{rashkin-etal-2017-truth}, we also extract the number of adjectives, the overall number of verbs and action verbs, as well as abstract nouns (e.g. ‘truth’, ‘freedom’), money symbols, assertive words, and second person pronouns (‘you’), and the first person singular (‘I’). Inspired by the journalistic rules of conduct for neutrality\footnote{ https://www.spj.org/ethicscode.asp}, we count the number of occurrences of words from several dictionaries: survey, reporting words, discourse markers, reflecting the surface coherence of the article, words denoting claims(e.g. ‘reckon’, ‘assume’), high modality words (e.g. ‘obviously’, ‘certainly’). As syntactic features, we measure the number of subordinate clauses of concession (using subordinate conjunctions similar to ‘although’, and ‘even if’), reason (e.g. ‘because’, ‘as’), purpose (e.g. ‘in order to’, ‘to’), condition (e.g. ‘provided’, ‘if’), time (e.g. ‘when’, ‘as soon as’) and relative clauses, which reflect different ways of justification and argumentation.

All features are automatically extracted in a pipeline and translated for each language using simplemma\footnote{https://adrien.barbaresi.eu/blog/simple-multilingual-lemmatizer-python.html} and pymorphy2\footnote{https://pymorphy2.readthedocs.io/en/stable/} for part-of-speech extraction in Ukrainian and spacy\footnote{https://spacy.io} for Russian, English, Romanian and French. The code will be released publicly upon acceptance.
\begin{table*}[!htbp]
\footnotesize
\centering
\caption{Comparative results achieved on best folds.}
\label{tab:results}
\begin{tabular}{|lllll|}
\hline
\rowcolor[HTML]{FFFFFF} 
\multicolumn{1}{|l|}{\cellcolor[HTML]{FFFFFF}Algorithm} &
  \multicolumn{1}{l|}{\cellcolor[HTML]{FFFFFF}Cohen's $\kappa$} &
  \multicolumn{1}{l|}{\cellcolor[HTML]{FFFFFF}F1} &
  \multicolumn{1}{l|}{\cellcolor[HTML]{FFFFFF}False positives} &
  False negatives \\ \hline
\rowcolor[HTML]{FFFFFF} 
\multicolumn{5}{|c|}{\cellcolor[HTML]{FFFFFF}Experiment 1.1 Test on subset (1768 texts)} \\ \hline
\rowcolor[HTML]{FFFFFF} 
\multicolumn{1}{|l|}{\cellcolor[HTML]{FFFFFF}Decision tree} &
  \multicolumn{1}{l|}{\cellcolor[HTML]{FFFFFF}0.49} &
  \multicolumn{1}{l|}{\cellcolor[HTML]{FFFFFF}0.73} &
  \multicolumn{1}{l|}{\cellcolor[HTML]{FFFFFF}16} &
  450 \\ \hline
\rowcolor[HTML]{FFFFFF} 
\multicolumn{1}{|l|}{\cellcolor[HTML]{FFFFFF}Linear logistic model} &
  \multicolumn{1}{l|}{\cellcolor[HTML]{FFFFFF}0.58} &
  \multicolumn{1}{l|}{\cellcolor[HTML]{FFFFFF}0.79} &
  \multicolumn{1}{l|}{\cellcolor[HTML]{FFFFFF}113} &
  265 \\ \hline
\rowcolor[HTML]{FFFFFF} 
\multicolumn{1}{|l|}{\cellcolor[HTML]{FFFFFF}SVM} &
  \multicolumn{1}{l|}{\cellcolor[HTML]{FFFFFF}0.75} &
  \multicolumn{1}{l|}{\cellcolor[HTML]{FFFFFF}0.87} &
  \multicolumn{1}{l|}{\cellcolor[HTML]{FFFFFF}156} &
  151 \\ \hline
\rowcolor[HTML]{FFFFFF} 
\multicolumn{1}{|l|}{\cellcolor[HTML]{FFFFFF}MLP} &
  \multicolumn{1}{l|}{\cellcolor[HTML]{FFFFFF}0.64} &
  \multicolumn{1}{l|}{\cellcolor[HTML]{FFFFFF}0.80} &
  \multicolumn{1}{l|}{\cellcolor[HTML]{FFFFFF}103} &
  229 \\ \hline
\rowcolor[HTML]{FFFFFF} 
\multicolumn{1}{|l|}{\cellcolor[HTML]{FFFFFF}BERT} &
  \multicolumn{1}{l|}{\cellcolor[HTML]{FFFFFF}0.84} &
  \multicolumn{1}{l|}{\cellcolor[HTML]{FFFFFF}0.92} &
  \multicolumn{1}{l|}{\cellcolor[HTML]{FFFFFF}97} &
  42 \\ \hline
\rowcolor[HTML]{FFFFFF} 
\multicolumn{5}{|c|}{\cellcolor[HTML]{FFFFFF}Experiment 1.2 Test on French (20 texts)} \\ \hline
\rowcolor[HTML]{FFFFFF} 
\multicolumn{1}{|l|}{\cellcolor[HTML]{FFFFFF}SVM} &
  \multicolumn{1}{l|}{\cellcolor[HTML]{FFFFFF}0.01} &
  \multicolumn{1}{l|}{\cellcolor[HTML]{FFFFFF}0.5} &
  \multicolumn{1}{l|}{\cellcolor[HTML]{FFFFFF}6} &
  16 \\ \hline
\rowcolor[HTML]{FFFFFF} 
\multicolumn{1}{|l|}{\cellcolor[HTML]{FFFFFF}BERT} &
  \multicolumn{1}{l|}{\cellcolor[HTML]{FFFFFF}0.05} &
  \multicolumn{1}{l|}{\cellcolor[HTML]{FFFFFF}0.52} &
  \multicolumn{1}{l|}{\cellcolor[HTML]{FFFFFF}19} &
  0 \\ \hline
\rowcolor[HTML]{FFFFFF} 
\multicolumn{5}{|c|}{\cellcolor[HTML]{FFFFFF}Experiment 2.1  Test on subset (1827 texts)} \\ \hline
\rowcolor[HTML]{FFFFFF} 
\multicolumn{1}{|l|}{\cellcolor[HTML]{FFFFFF}SVM} &
  \multicolumn{1}{l|}{\cellcolor[HTML]{FFFFFF}0.75} &
  \multicolumn{1}{l|}{\cellcolor[HTML]{FFFFFF}0.884} &
  \multicolumn{1}{l|}{\cellcolor[HTML]{FFFFFF}120} &
  \cellcolor[HTML]{FFFFFF}151 \\ \hline
\rowcolor[HTML]{FFFFFF} 
\multicolumn{1}{|l|}{\cellcolor[HTML]{FFFFFF}BERT} &
  \multicolumn{1}{l|}{\cellcolor[HTML]{FFFFFF}0.86} &
  \multicolumn{1}{l|}{\cellcolor[HTML]{FFFFFF}0.93} &
  \multicolumn{1}{l|}{\cellcolor[HTML]{FFFFFF}111} &
  12 \\ \hline
\rowcolor[HTML]{FFFFFF} 
\multicolumn{5}{|c|}{\cellcolor[HTML]{FFFFFF}Experiment 2.2 Test on Telegram (14525 texts)} \\ \hline
\rowcolor[HTML]{FFFFFF} 
\multicolumn{1}{|l|}{\cellcolor[HTML]{FFFFFF}SVM} &
  \multicolumn{1}{l|}{\cellcolor[HTML]{FFFFFF}0.25} &
  \multicolumn{1}{l|}{\cellcolor[HTML]{FFFFFF}0.637} &
  \multicolumn{1}{l|}{\cellcolor[HTML]{FFFFFF}2013} &
  3402 \\ \hline
\rowcolor[HTML]{FFFFFF} 
\multicolumn{1}{|l|}{\cellcolor[HTML]{FFFFFF}BERT} &
  \multicolumn{1}{l|}{\cellcolor[HTML]{FFFFFF}0.17} &
  \multicolumn{1}{l|}{\cellcolor[HTML]{FFFFFF}0.58} &
  \multicolumn{1}{l|}{\cellcolor[HTML]{FFFFFF}5770} &
  212 \\ \hline
\rowcolor[HTML]{FFFFFF} 
\multicolumn{5}{|c|}{\cellcolor[HTML]{FFFFFF}Experiment 3. Test on subset (8709 texts)} \\ \hline
\rowcolor[HTML]{FFFFFF} 
\multicolumn{1}{|l|}{\cellcolor[HTML]{FFFFFF}SVM} &
  \multicolumn{1}{l|}{\cellcolor[HTML]{FFFFFF}0.66} &
  \multicolumn{1}{l|}{\cellcolor[HTML]{FFFFFF}0.88} &
  \multicolumn{1}{l|}{\cellcolor[HTML]{FFFFFF}707} &
  267 \\ \hline
\rowcolor[HTML]{FFFFFF} 
\multicolumn{1}{|l|}{\cellcolor[HTML]{FFFFFF}BERT} &
  \multicolumn{1}{l|}{\cellcolor[HTML]{FFFFFF}0.81} &
  \multicolumn{1}{l|}{\cellcolor[HTML]{FFFFFF}0.92} &
  \multicolumn{1}{l|}{\cellcolor[HTML]{FFFFFF}136} &
  162 \\ \hline
\end{tabular}
\end{table*}
\subsection{Keywords}
As a list of keywords, we use the glossary\footnote{https://www.rnbo.gov.ua/files/2021} prepared by the National Security and Defense Council of Ukraine. It contains a list of names, terms and phrases recommended for use by public authorities and diplomatic missions of Ukraine and as well as versions of these terms used in Russian political discourse.
We translate this glossary to the target languages and add a short list from the military lexicon being avoided by Russian officials (e.g. 'war', 'victims', 'children', 'casualties') \cite{mention}.

\section{Results}\label{sec2}
\label{sec:results}

We evaluate the performance of our models using Cohen's $\kappa$ \cite{cohen} and F-measure \cite{f1}.
While F1-score is easy to interpret and most frequently used, subtracting the Expected Accuracy, Cohen’s Kappa removes the intrinsic dissimilarities of different data sets, which makes two classification problems comparable, as K can compare the performances of two models on two different cases \cite{Grandini2020MetricsFM}.
We also evaluate the number of false positives and negatives, which help build a complete picture of the model's performance. The results for all settings, averaged over five models, can be found in Table \ref{tab:results}. Details about the models and hyperparameters can be found in Appendix.

\textbf{Experiment 1.} When training on Russian, Ukrainian, Romanian and English newspaper articles, the best result on the handcrafted linguistic features (no keywords) was achieved with an SVM: 0.87 F1-measure and 0.75 $\kappa$. The model is almost equally prone to false positives (108) and false negatives (120) across 1768 test samples (FP-rate: 0.06, FN-rate: 0.06). Linear models and a 2-layer neural network performed considerably worse (F1: 0.8). As the SVM performed best, we continued our experiments with this model, added our extracted keywords to the dataset, but found no improvement. 

The linear model using BERT embeddings outperformed the handcrafted feature models (F1: 0.92, and $\kappa$: 0.84). While it produces a similar quantity of false positives as the SVM, the false negative rate decreases considerably.\\
When testing on 40 French texts (20 pro-Kremlin, 20 pro-Ukrainian), the performance drops considerably for the feature-based approach (F1: 0.5, $\kappa$: 0.01) with 14 false negatives and 6 false positives, and for BERT embeddings (F1: 0.52, $\kappa$: 0.05) with 19 false positives and only one true negative.

\textbf{Experiment 2.} The addition of French newspaper articles to the training set increased the F1-score by 0.08 for both SVM and embeddings-based models. However, the models do not perform well when tested on Telegram data. Without keywords, the SVM model scored 0.61 F1-measure, with a very low $\kappa$ of 0.24, 2078 false positives and 3422 false negatives out of 14525 test samples. Adding keywords increases performance (F1: 0.62, $\kappa$: 0.25), lowering the false positive and false negative
(FP-rate: 0.13, FN-rate: 0.23).
The embeddings-based model scores even lower (F1: 0.58, $\kappa$: 0.17, FP-rate: 0.39, FN-rate: 0.014)

\textbf{Experiment 3.} Finally, we train on the full dataset with both newspaper articles and Telegram posts. The handcrafted feature-based model increases the F1-score to 0.88, but decreases $\kappa$ to 0.66, with 707 false positives and 267 false negatives out of 8709 test samples. The embeddings-based model reaches 0.90 F1-measure and 0.81 $\kappa$, with 136 false positives and 162 false negatives.\\
Both models make disproportionately more errors when tasked with the classification of Romanian texts.

\begin{figure*}[!htbp]
\centering
\includegraphics[width=16cm,height=6cm]{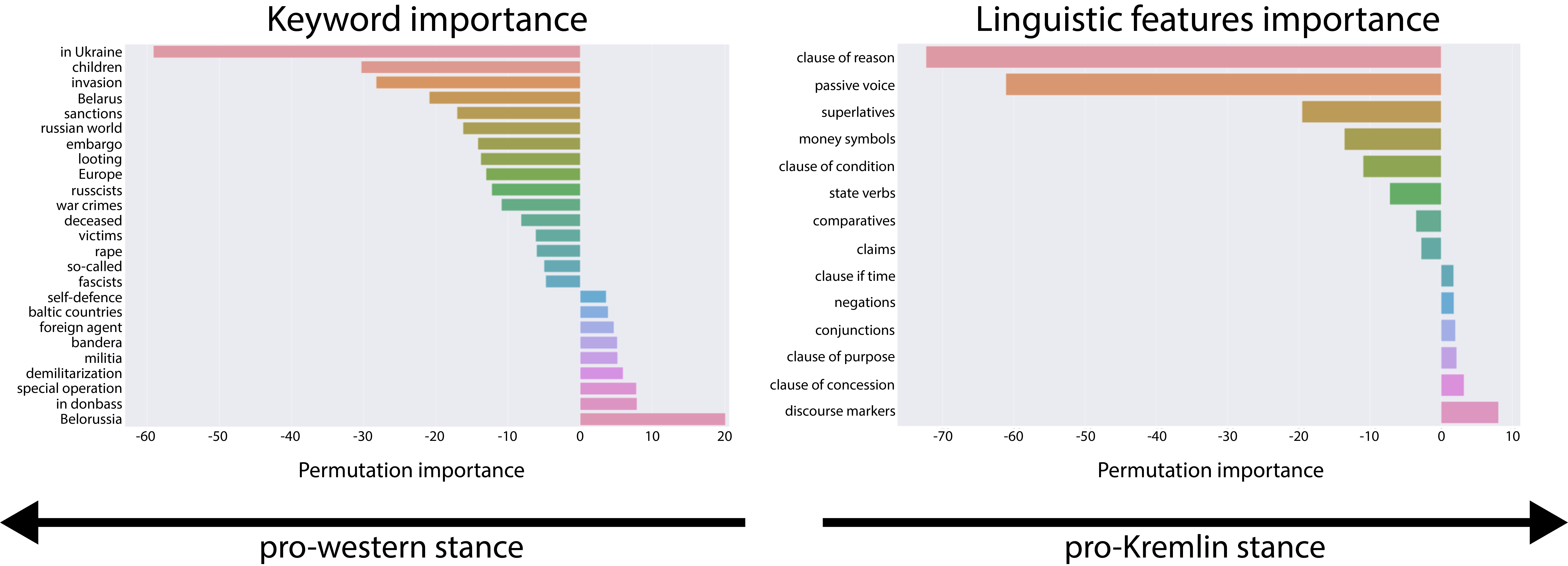}
\caption{Permutation importance (drop of F1 score in \%) for an SVM with linear kernel. Keyword importance is on the left side and the importance of linguistic features is on the right. Negative bars indicate features that are important for classifying a data point as pro-Western, while positive bars represent features indicative of pro-Kremlin propaganda. }\label{weights}
\end{figure*}

\section{Additional Analysis}
\label{sec:analyses}

\subsection{Feature Importance}

We further analyse our best-performing SVM model to obtain feature importance for both linguistic features and keywords using the feature permutation method \cite{Breiman2004RandomF}.
We find that various subordinate clauses prove to be important for the model, with the presence of the clause of the reason being the most indicative of pro-Western narratives, as well as passive voice. To a lesser degree, the following features were also deemed as important: superlatives, money symbols and words, clauses of condition, state verbs, comparatives and words indicating claims. For those features, it can be stated that they are unlikely to be found in pro-Kremlin news.
At the same time, discourse markers (e.g. ‘however’, ‘finally’) as well as clauses of concession, clauses of purpose, conjunctions, negations and clauses of time separate pro-Kremlin news the best.

We find many keywords coming from the list provided by the Ukrainian Security Council glossary to be important. However, some of them need cultural and historical context to be understood. We find that the formulation ‘in Ukraine’ is the most reliable marker of pro-Western news, while in Russian news the formulation is ‘on Ukraine’, which indicates its use as ‘on a territory’ and not ‘in the country’. Interestingly, the use of ‘in Donbas’ is the second highest indicator for Russian news. While it is a conventional name for the territory shared by two Ukrainian regions, it would preferably be used with the preposition ‘on’, e.g. ‘on the Western Balkans’. The usage of 'in' gives linguistic legitimacy to the idea of the independence of the quasi-republics. The variant of the country's name ‘Belarus’ is highly indicative of the Western side, while ‘Belorussia’, the version found in Russian news, presents the neighbouring country’s name rather as ‘white’ Russia, and not as ‘Rus’, the historical area in Eastern Europe. The formulation ‘special operation’\footnote{https://www.un.org/press/en/2022/sc14803.doc.htm} is a euphemism for the word ‘war’ used by the Russian government and the pro-governmental news agencies. It is a strong indicator towards a  pro-Kremlin narrative. On the Western side, we observe that the word ‘invasion’ has a higher frequency. Other words with high importance values for pro-Kremlin narratives are demilitarisation (of Ukraine), ‘self-defence’, ‘militia’, ‘Bandera’ \footnote{Politician and theorist of the militant wing of the Organization of Ukrainian Nationalists in 20th-century \cite{marples2006stepan}.}, ‘Baltic countries', also commonly called ‘Pribaltika’ in Russian, again presented more as a territory, and finally ‘foreign agent’.

Many of the words we assumed will not be used in pro-governmental Russian articles were found to be important markers. Hence, words commonly used by the pro-Ukrainian news describing the disastrous consequences of war for both sides, e.g. ‘children’, ‘looting’, ‘war crimes’, ‘deceased’, ‘victims’, ‘rape’, ‘sanctions’, ’embargo’, are attributed high importance in Western media. Some other curious keywords often occurring in Western media are ‘Russian world’ and ‘russcists’, which are mainly used by Ukrainian media as means of referring to the ideology of the Russian military expansionism \cite{gaufman2016security}.

\subsection{Distributional exploratory analysis}

\textbf{Chronological analysis.} We also carried out an exploratory study of the feature and keyword distributions over 5 data collections: the 23rd of February, the 1st of March, the eighth of March, the 18th of March and the fourth of April. We looked at the contrast between different languages and between pro-Kremlin and pro-Western media within one language, with the aim to explain frequent model errors and observe how media reflects the events of the war.

The most noticeable observation for Ukrainian pro-Western media is an increase in many features on the 18th of March, following the bombing of the Mariupol theatre on the 16th \cite{mariupol}: abstract nouns, claims, and negative emotional lexicon (namely words of surprise, disgust, sadness, fear and anger). Some indicators, like reporting words, negations, proper nouns, and modal verbs drop in frequency in March and seem to come back to the pre-war level in April. The use of the word war is constantly increasing throughout our data collection.

In contrast, in Russian pro-governmental media, the collection date with the most deviation from the overall average is the 1st of March, when we can observe a drastic increase in the number of adjectives, average sentence length, assertive words, clauses of purpose, but also negative emotional lexicon (including words of trust and anger), and positive emotional lexicon. 1st of March corresponds to the end of the first week of the war when it became clear that the war might extend for a longer period \cite{kriegplans1}.

British and American media remained quite stable throughout this time period, although we can observe an increase in superlatives on the fourth of April, which follows closely the discovery of the war crimes in Bucha, where 412 civilians were killed \cite{bucha}. Pro-Western Romanian data also did not change considerably, with an exception of a slight increase with each collection in clauses of reason, words of surprise and the keyword ‘war’. At the same time, in the Romanian media flagged as fake news, there is a drop in words of anger and an increase in words of disgust, surprise, happiness and expectation, as well as abstract nouns, modal verbs, clauses of purpose when compared to the pre-war collection.


\textbf{Language and narrative specific feature differences.} When comparing media in different languages we observe interesting trends, which, however, did not account much for the decision of the classifiers. For instance, English and Ukrainian pro-Western media have the highest personal pronouns frequency, while newspapers from Russian media have the highest amount of quotations and are the only using keywords such as: ‘coup d'etat’, ‘DNR’, ‘LNR’, ‘Kiev regime’, ‘russophobia’.
Romanian articles from trusted sources have the longest sentences, and all articles from Romanian media have the lowest use of the conjunction ‘but’. Furthermore, all articles have the highest occurrence of comparatives, superlatives and state verbs, which we believe is language specific. This might be the reason for the low performance when applied to Romanian articles since these three features have high importance for the SVM model.

Articles from Ukrainian media generally have a high frequency of adjectives. At the same time, Ukrainian pro-Western news has the highest amount of emotional words (sadness, expectation, disgust, surprise, fear, anger), while pro-Russian Ukrainian articles do not show such a tendency, and thus might not reflect the same emotional atmosphere. It also has the same distribution of clauses of time as Russian pro-governmental news and the equally lowest usage of passive verbs.

Thus, we do not see a clear tendency for Russian propaganda to be homogeneous among the countries we selected. The only example would be the use of the keyword ‘Soros’, which is used uniquely in pro-Kremlin media in Ukraine and Russia, as well as in fake-news-flagged Romanian media. This can be explained by invocations of antisemitic conspiracy theories explored in Timothy Snyder's ‘Road to Unfreedom’\cite{ nla.cat-vn7732877} as manipulation strategies. Otherwise, media in Romania seems to be much more adapted to the local journalistic specifics, while in Ukraine the pro-Kremlin articles have much more in common with their origin.

\section{Discussion}
\label{sec:disussion}

Our study is the first attempt to quantitatively and qualitatively analyse and automatically detect Russian propaganda, which does not necessarily use an emotional lexicon. Surprisingly, the propaganda style seemingly mimics international codes of conduct for journalism and adapts to each target language and it is country-specific. Many features of the Western news class can be found in the aforementioned related works for fake news detection, while pro-Kremlin features taken out of context could be interpreted as neutral journalistic style. This indicates that morpho-syntax modelling and analysis without semantics may be misleading. Both sides use propaganda. We found that their features differ, which may be explained by different ideologies and cultural specifics, but it may indicate different goals.
Russian-backed media justifies the war and downplays adverse effects, while Ukrainian war propaganda focuses on various emotions, from ridiculing Russian forces, instigating hate against the Russian people as a whole or proposing an over-optimistic view of the military situation.
In future work, we propose including an additional class representing neutral Ukrainian, Russian, and international news. However, labelling such datasets would require much more time for manual annotation.\\
Since the appearance of state-of-the-art Transformer-based models, the trade-off between transparency and accuracy is a topical issue in the NLP community. We show that transparent statistical models based on linguistic expert knowledge can still be competitive. Our best embeddings-based model has only around 0.04 F1-score advantage, but it is less explainable. We cannot control if the BERT model learnt the style of the media outlets instead of propaganda itself, while we can be sure that SVM indeed captures Kremlin's manipulative lexicon. While there are methods to interpret decisions \cite{interpret} of such models, we leave this for future work. As BERT models can capture syntax \cite{jawahar-etal-2019-bert}, we believe that such embeddings might still be less flexible towards changes in themes and topics, and need retraining if major vocabulary changes occur. The BERT-based model has a clear tendency towards false positives and performs slightly worse on the data from different distributions. In the context of automated content moderation, false positives would mean flagging/filtering a post or banning a person, limiting the freedom of speech. False negatives might lead to posts with propaganda reaching more targets.

Keywords were only beneficial for SVM when applied to new data, where the algorithm had to base its decisions on semantics more than morpho-syntax. The overall journalistic style captured by handcrafted features is more reliable, as the model performance does not drastically change for any of the languages in focus, even in the face of such events as war. Scalability is, however, a major drawback of feature-based models, as new predictions require first-feature extraction, while models using BERT embeddings can be used out of the box. 
However, BERT models have important token length limitations, whereas with SVM we pass a stable vector of feature counts normalised by the text length. 
While it might seem natural to choose these high-performing models in industrial settings, we believe that for the sake of transparency, models using handcrafted features that are competitive can still be used.\\
Both approaches can turn out to be inefficient after a certain period of time, especially in light of the new tendencies towards automatically generated news.

We see our work as a first step towards a browser extension flagging harmful content to raise individual awareness and assist us in filtering the content we consume. However, the classifier can be used to block pro-Western news as well, ensuring the impenetrability of echo chambers, amplifying the effects of propaganda instead of helping to fight it.  

\begin{figure}
\includegraphics[width=8cm,height=4cm]{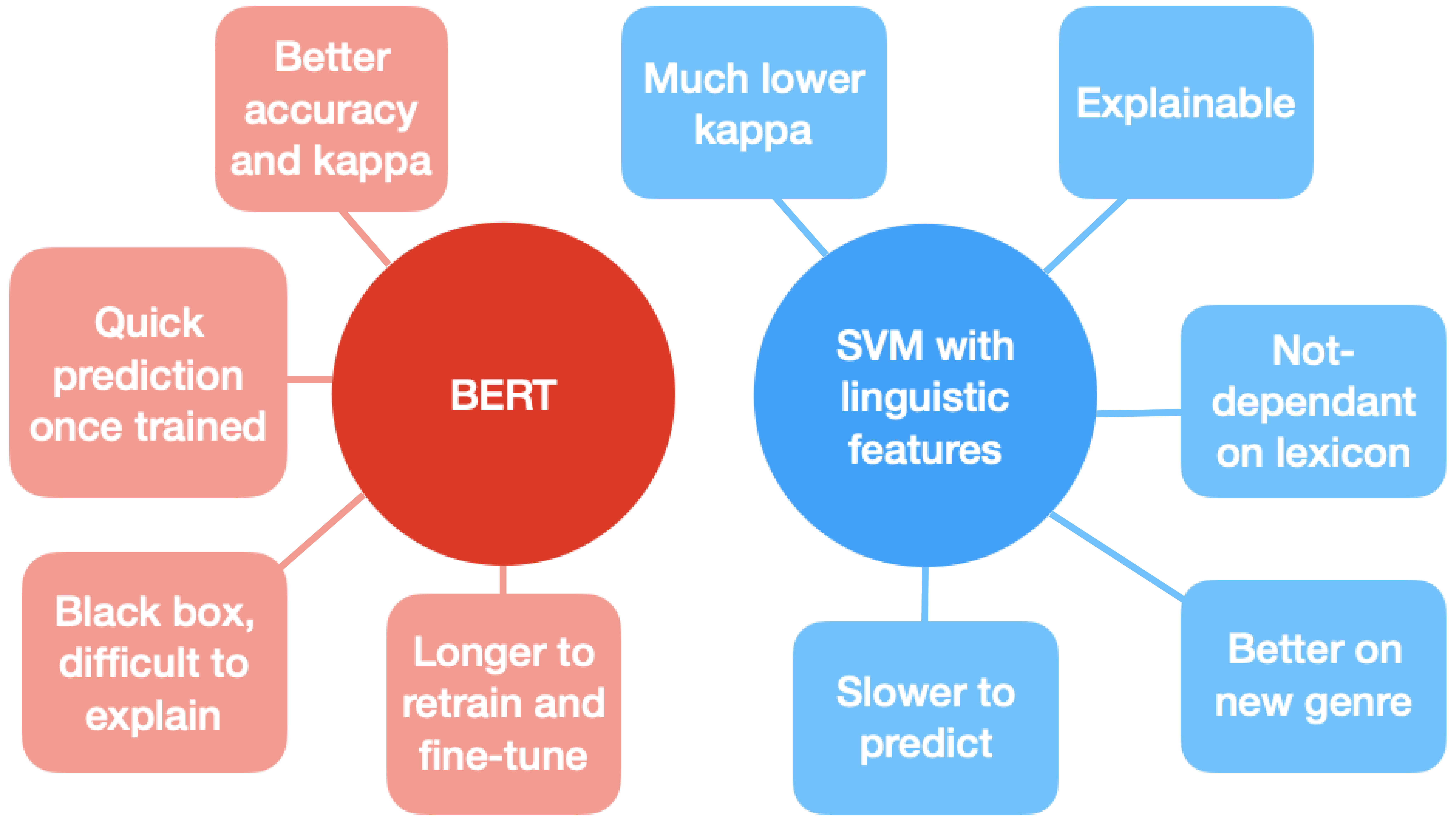}
\caption{Advantages and disadvantages of the presented methods.}
\label{proscons}
\end{figure}

\section{Conclusion}\label{sec13}

We presented two alternative methods to automatically identify pro-Kremlin propaganda in newspaper articles and Telegram posts. Our analysis indicates that there are strong similarities in terms of rhetoric strategies in the pro-Kremlin media in both Ukraine and Russia. While being relatively neutral according to surface structure, pro-Kremlin sources use artificially modified vocabulary to reshape important geopolitical notions. They also have, to a lesser degree, similarities with the Romanian news flagged as fake news, suggesting that propaganda may be adapted to each country and language in particular. Both Ukrainian and Russian sources lean towards strongly opinionated news, pointing towards the use of war propaganda in order to achieve strategic goals.

Russian, Romanian and Ukrainian languages are under-researched in terms of NLP tools in comparison to English. We hope that our study contributes to social media and individual efforts to moderate harmful content. We share our data and code as open-source tools for the detection of automated fake news or propaganda in order to help local human moderators and common users in those countries.
\section{Acknowledgements}\label{sec14}
The research was completed while the second author was still affiliated with Freie Universität Berlin.

\backmatter

\section*{Declarations}

\begin{itemize}
\item Funding \\
Not applicable
\item Competing interests \newline 
The corresponding author VS is of Ukrainian nationality but is not affiliated with any of the Ukrainian governmental or private institutions. Other authors have no competing interests to declare.
\item Ethics approval\\
Not applicable
\item Consent to participate\\
Not applicable
\item Consent for publication\\
Not applicable
\item Availability of data and materials\newline
All data is available in a fully anonymous until publication GitHub depository\footnote{https://github.com/anonrep/pro-kremlin\_propaganda}.
\item Code availability \\
All code is available in the same GitHub repository.
\item Authors' contributions\\
 Conceptualization: [Veronika Solopova]; Methodology: [Veronika Solopova, Oana-Iuliana Popescu]; Formal analysis and investigation: [Veronika Solopova, Oana-Iuliana Popescu]; Writing - original draft preparation: [Veronika Solopova]; Writing - review and editing: [Veronika Solopova, Tim Landgraf, Oana-Iuliana Popescu]; Supervision: [Tim Landgraf, Christoph Benzmüller].
\end{itemize}


\begin{appendices}

\section{Hyper-parameters}\label{secA1}

We performed a Grid search and found out that the best results are achieved with Radial basis function kernel, gamma=100, and C=46 parameters.\\ \\
For the neural network used with linguistic features, our best setup was achieved with 2 hidden layers, a Limited-memory BFGS solver, tanh activation function, and alpha=1e-5.\\ \\
For the linear model used with BERT, we used a learning rate of 1e-4, 4 epochs and batch size 16.

\end{appendices}
\providecommand{\latin}[1]{#1}
\makeatletter
\providecommand{\doi}
  {\begingroup\let\do\@makeother\dospecials
  \catcode`\{=1 \catcode`\}=2 \doi@aux}
\providecommand{\doi@aux}[1]{\endgroup\texttt{#1}}
\makeatother
\providecommand*\mcitethebibliography{\thebibliography}
\csname @ifundefined\endcsname{endmcitethebibliography}
  {\let\endmcitethebibliography\endthebibliography}{}



\end{document}